# Zipf's law in 50 languages:[*]

## its structural pattern, linguistic interpretation, and cognitive motivation

Shuiyuan YU[1], Chunshan Xu[2, 3], Haitao LIU[3, 4]

1 School of Computer, Communication University of China, Beijing,100024, China
2 School of Foreign Studies, Anhui Jianzhu University, Hefei, 230601, China.
3 Department of Linguistics, Zhejiang University, Hangzhou, 310058, China
4 Ningbo Institute of Technology, Zhejiang University, Ningbo, 315100, China

**Abstract**   Zipf's law has been found in many human-related fields, including language, where the frequency of a word is persistently found as a power law function of its frequency rank, known as Zipf's law. However, there is much dispute whether it is a universal law or a statistical artifact, and little is known about what mechanisms may have shaped it. To answer these questions, this study conducted a large scale cross language investigation into Zipf's law. The statistical results show that Zipf's laws in 50 languages all share a 3-segment structural pattern, with each segment demonstrating distinctive linguistic properties and the lower segment invariably bending downwards to deviate from theoretical expectation. This finding indicates that this deviation is a fundamental and universal feature of word frequency distributions in natural languages, not the statistical error of low frequency words.  A computer simulation based on the dual-process theory yields Zipf's law with the same structural pattern, suggesting that Zipf's law of natural languages are motivated by common cognitive mechanisms. These results show that Zipf's law in languages is motivated by cognitive mechanisms like dual-processing that govern human verbal behaviors.

---

**Introduction**

About 80 years ago, George Kingsley Zipf reported an observation that the frequency of a word seems to be a power law function of its frequency rank, formulated as f(r) $\propto r^{\alpha}$, where *f* is word frequency, *r* is the rank of frequency, and *α* is the exponent(*1, 2*). This linguistic regularity was later termed as Zipf's law. In subsequent years, similar power law functions have been widely found in various domains, drawing attention of scientists with different academic backgrounds—it may reflect a universal law underlying various natural, social and cognitive phenomena (*3-5*). To specify the mechanisms motivating such power-law functions, much work has been done from different perspectives (*6-19*). However, a universally acknowledged conclusion is yet to obtain, which reflects a failure to fully understand the role and the significance of these power functions in natural, social and cognitive spheres.

In an effort to explore the motivations of Zipf's laws, we conducted a case study on natural languages. Though this law was first observed in language, its significance in verbal communication and deep-level motivations are far from being clear. We know neither why this law has been persistently found in samples of different languages, and with different sizes, nor what underlying mechanisms may have shaped it. As a result, it is often proposed that this law has no linguistic significance and cognitive motivations: simple and meaningless statistical processes (*16*), such as random typing model (*6, 8, 20*), can result in power-law distributions similar to Zipf's law as well.

However, a law usually describes the fundamental nature of something. The failure to pin down deep level motivations actually suggests that Zipf's law is not a law: it sheds no light on the fundamental nature of verbal communication and the cognitive mechanisms underlying it. To trace the possible cognitive mechanism that shapes this law, it is necessary to find a universally valid linguistic interpretation of Zipf's law, or rather its universal significance in verbal communication, which means that Zipf's curve (word frequency as a function of rank on a log-log scale) should exhibit, across different languages, similar structural patterns that have similar linguistic significance. Hence, corpus-based investigations are needed that cover languages as diverse as possible. Nevertheless, previous studies were generally limited to a rather limited range of languages and paid little attention to its underlying cognitive mechanism and significance in language communication.

Based on texts from 50 languages, this quantitative study reveals that Zipf's law found in these languages all exhibit a 3-segment structural pattern, with the lower segment invariably bending downward to deviate from the theoretical expectation, which has linguistic significance and can be construed as motivated by dual-process model—a cognitive mechanism common among human beings. We further demonstrate that computer simulation based on this cognitive mechanism can generate Zipf's curves that are structurally similar to those found in human languages. These findings suggest that Zipf's laws found in human-related fields are likely to have deep-level motivations, especially the cognitive ones, and that big-data analysis based on corpus may provide valuable means to uncover language laws and to trace cognitive laws hidden behind.

**Materials and methods**

**Data**

Language materials used in our study are taken from BNC and Leipzig Corpus. In these corpora, we choose 50 languages. The sample of each language contains over 300,000 sentences, ranging between 80,000 and 43,000,000 word tokens. These languages belong to such typological branches as Indo-European, Uralic, Altaic, Caucasian, Sino-Tibet, Dravidian, Afro-asiatic, etc. Of these 50 languages, 25 include texts of both news and wiki.

To avoid the stochastic influence of text genres on overall word frequency, we conduct text-based randomization upon the materials from BNC. The Leipzig Corpus has already been randomized on the basis of sentences.

**Piecewise fitting of frequency-rank curves of 50 languages to different power-law functions**

The rank-frequency curves found in these 50 languages seem all to exhibit a 3-segment structural pattern. We therefore fit these 3 segments respectively to the following power law functions: $y = a \cdot x^b$, $y = a \cdot x^b$, and $y = a \cdot (x + c)^b$. In other words, for the fitting function of the lower segment that covers the words with low frequency, the definition domain is not solely confined to the frequency rank. In addition, to prevent the influence of too many words with extremely low frequency, the fitting of the lower segment is based on logarithmically equal size spaces, not linearly equal size space, and offset C is added to the fitting function to indicate the steepness and the direction of the bending.

**The influence of text genres on the 3 segments of the curves**

25 languages are chosen in Leipzig corpus to find out the influence of the genres on the exponents of fitting functions of these curves. For each language, the sample covers two genres, including both wiki texts and news texts. For each genre, there are more than 300, 000 sentences. The exponents derived from texts of wiki are paired with the exponents derived from texts of news to conduct Wilcoxon rank sum test. If the P-value of test is smaller than alpha, the null hypothesis is rejected that parameters of fitting function are not influenced by genres.

**Relations between linguistic typologies and function exponents of each curve segment.** To study the effect of typological differences, we chose, among the 50 languages, 37 languages that fall into 7 typological branches that each contains at least 3 of these 37 languages. In other words, these 37 languages are unevenly divided into 7 groups to conduct a F-test, which can reveal whether typological differences affect exponents of fitting functions.

**The growth rate of word frequency in upper segment, middle segment, and lower segment.** To investigate how word frequency grows with increasing sample size, the following formula is used to calculate the average growth rate of word frequency: $\frac{1}{N-1}\sum_{n=1}^{N-1}\frac{f(n+1)/f(n)}{c(n+1)/c(n)}$, in which f(n) is the frequency of a word in language sample whose size is n word tokens. For each language, $S_1 = S_2 = S_3 = \ldots\ldots S_n$ are samples with different sizes; N(S) is the number of word tokens in a sample; the relations among these samples are $S_1 \subset S_2 \subset S_3 \subset, \ldots\ldots \subset S_n$ and 2 $C(S_1) = C(S_2), 2C(S_2) = C(S_3), \ldots\ldots 2C(S_{n-1}) = C(S_n)$.

We also investigate the relation between the number of word types and the number of word tokens (sample size) through linearly fitting the logarithm of number of word types to the logarithm of the number of word tokens of each sample in a language. The monomial coefficient of the fitting function is the Heaps' exponent. To examine the influence of genre and typology on Heaps' exponent, a F-tests are conducted on the news texts and wiki texts of 25 languages, and on the samples of 37 languages that fall into 7 typological branches.

**Computer simulations of dual-generating mechanism model .**This model includes two mechanisms to simulate the growth of word tokens, one for the 3000 high-frequency words that are covered in the upper and the middle segments of the curve, and the other for low-frequency words and new words that are covered in the lower segment of the curve. For high-frequency words, the increase of sample size will cause their total frequency to escalate according to a fixed probability as found in existent samples. In other words, total probability of high frequency words (their relative frequency) is not much influenced by the variation in sample size. The remainder probability (1 minus the total probability of high frequency words) is assigned to low frequency words and new words. For example, if the total probability of high frequency words is 0.8, the frequency of these words will stably increase in accordance with the increase of sample size, and thus the total probability will remain steadily as about 0.8. At the same time, the remainder 0.2 probability will be assigned to low frequency words. Since the total probability is persistently 0.2, more new words that may appear in a larger sample will reduce the average occurrence probability (or the mean relative frequency) of each word, which shapes the downward bending of the lower segment.

**Results**

Zipf's law reflects the power law relation between word frequency and its rank. Since Zipf's law is found, in our study, invariably to exhibit a 3-segment structural pattern, we are mainly concerned with dynamic patterns and the limiting values of the power exponents of these 3 segments, that is, the power exponent of the upper segment (hereafter exponent 1), the power exponent of the middle segment (hereafter exponent 2), and the power exponent of the lower segment (hereafter exponent 3). The results of fitting variations of exponent 1 and 2 in samples of increasing size to power law, with adjusted $R^2$=0.8506 and 0.9177, show that, when the sample size approaches infinity, increasing sample size results in zero difference of exponent 1 and 2 go to, and that exponent 3 has a limiting value, with the average, the maximal, and the minimal goodness of fitting being respectively 0.9603, 0.5277, and 0.9996. In other words, with increasing sample size, exponents 1 and 2 gradually reach constants, exponent 3 goes to a limiting value. Since exponent 3 is always significantly smaller than exponents 1 and 2, the downward deviation of the lower is a universal phenomenon in natural languages.

As seen in Figure 1, for all the 50 languages, the frequency-rank curves all deviate from the theoretical expectation of Zipf's law. For any of these 50 languages, the Zipf's curve can be dissected into 3 segments. Piecewise fitting of Zipf's curves to power-law functions in 50 languages indicates that the minimal adjusted $R^2$s of the 3 segments are 0.9619, 0.9908, and 0.9482; the maximal adjusted $R^2$s are 0.9976, 0.9998, and 0.9966; the mean adjusted $R^2$s are 0.9879, 0.9976, and 0.9831.The upper segment of the curve is consistently unsmooth, roughly covering words whose frequency ranks are within 200 (the 200 most frequent words); the middle segment is smoother, covering the words whose frequency ranks are between 200 and 2000 (or 3000 in some languages); the lower segment is also smooth, but slopes more steeply (that is, this segment bends downward), covering the rest of the words in the sample.

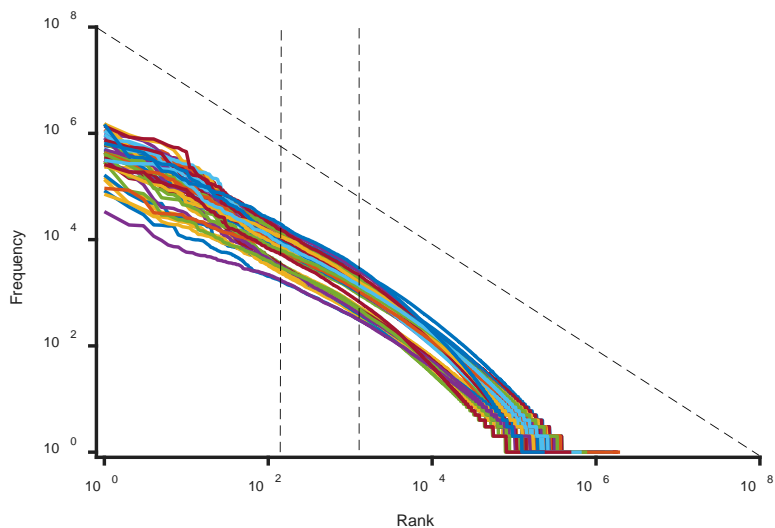

Figure 1．The frequency-rank curves found in 50 languages of Leipzig corpus. The upper segment of each curve (the segment before the 1st vertical line) is unsmooth; the middle segment of each curve (the segment between the 1st and the 2nd vertical lines) is smooth, with the gradient being roughly 1. The lower segment of each curve (the segment after the 2nd vertical line) is also smooth, but bends downward to deviate from the expected line.

In addition, the 3 segments also seem to register distinctive dynamic properties. As can be seen in Figure 2, the sample size has different influences on the 3 segments: 1) exponent 1 and exponent 2 remain stable despite the growth of sample size; 2) exponent 3 and the power exponent of the entire curve decrease with the growth of sample size, going to a limiting value when the sample size is approaching infinity. In other words, these 3 segments exhibit different dynamic properties. Due to limited space, Figure 2 demonstrates only the statistics of English. In fact, for the exponents of the 3 segments in 50 languages, the Pearson's linear correlation coefficients are 0.2543 between exponent 1 and 2, 0.0746 between exponent 1 and 3, -0.2384 between exponent 2 and 3. So Zipf's curves of 50 languages all show similar patterns. And statistical tests indicate no correlation between exponents of any two segments of the curves.

Apart from the sample size, linguistic factors, such as text genre and typological difference, seem to also have different effects on the 3 segments. Thus, 25 languages with both wiki texts and news texts are chosen to study the influence of genres on the 3 segments. On the basis of these 25 languages, this study, first of all, investigated the Ochiai coefficients of words that are shared by texts of both genres in each segment. The mean values are respectively 0.6362, 0.524, and 0.3477 in the 3 segments. In other words, in the upper segment, about 36% of words are different across the two genres, while for the lower segment, more than 65% of words are different. Such a difference may have much to do the different topics that different text genres usually involve. Then, we use nonparametric tests to investigate the influence of genres on exponents 1, 2, and 3, since little is known of the distribution patterns of the exponents of theses languages. A Wilcoxon rank sum test is conducted to investigate the influence of genres on the exponents of each segment, which indicates that the P-values of exponent 1, 2, 3 and Heaps' are respectively 0.3033, 0.0598, 0.3417, and 0.1160. Obviously, the test only yields low p-value for exponent 2. In other words, if we set significance level as 0.1, exponent 1, 3 and Heaps' are insensitive to genre differences while exponent 2 is sensitive. To investigate the influence of typological differences on exponents 1, 2, 3 and Heaps' exponent, we conduct F-test on 40 languages that fall into 7 typological branches. The results indicate that the F-values and P-values of Exponent 1, 2, 3, and Heaps' exponent are 22.98 and 1.7786e-10, 13.36 and 1.26e-07, 0.8742 and 0.5243, 0.7701 and 0.5988, which suggests

significant influences of typological difference on exponents 1 and 2. . Therefore, it can be seen that genre differences influence exponent 2, and typological differences influence both exponents 1 and 2. For the upper segment of the curves, most words remain the same regardless of different text genres. For the lower segment, on the contrary, most words are different across different genres. However, the exponents of segment 1 and 3 are insensitive to genres, suggesting that semantics has little influence on the overall frequency distributions of words in these two segments. What is more, these segments seem to cover words of different categories: the upper segment mainly includes the function words while the middle and the lower segment mainly cover content words.

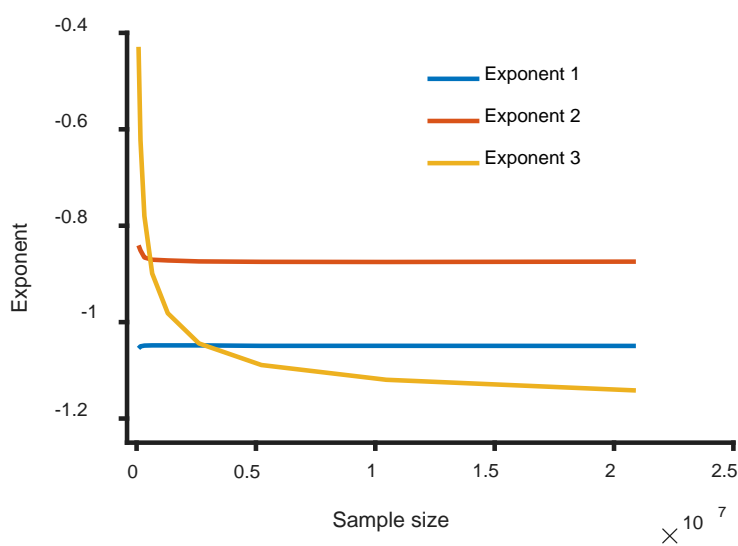

Figure 2．The relations between exponents 1, 2, 3 and the sample size as found in English sample of Leipzig corpus

The different dynamic properties of the 3 segments can be seen more clearly in the fact that, for different segments, the increase of sample size will bring about different growth rates of word frequency. The statistics of 50 languages indicate that the frequency of most words in the upper and the middle segments increase in proportion to the sample size, while the frequency of words in the lower segment does not. In other words, the upper and the middle segments of curves rise in accordance with the growth of the sample size, but the lower segment rises much slower, leading to the downward bending of the curve.

For large scale language samples, especially those that have been shuffled, the probability of words occurring in it is expected to be stable, and as a result, their frequency is expected to be in a constant proportion to sample size (the number of word tokens). To investigate the growth rates (dynamic properties) of word frequency in each segment, we calculate the growth rates of word frequency in relation to doubled sample sizes. The results show that, for most words in the upper and the middle segments, the growth rates cluster around 2 when sample size doubles. That is, frequencies of most words grow in proportion with the increase of sample sizes. In contrary, for most words in the lower segment, the growth rates cluster around 1, which means that, in this segment, the frequencies of most words grow much slower than the increase of sample sizes. Since the frequencies of most words in the lower segment fail to increase proportionally with the growth of the sample size, the relative frequency of words in this segment is likely to diminish in larger samples, or rather, the temporal interval between two occurrences of a word is likely to widen in larger samples.

In short, word frequency distributions in natural languages are characterized by a tendency for Zipf's curves of larger samples to bend increasingly downward. This phenomenon can be explained in terms of the cognitive mechanism of the dual-process model (*21-25*) that suggests two quite different generating mechanisms for high-frequency words and low-frequency words. To test this hypothesis, we conduct a computer simulation to find out whether a model including two different generating mechanisms is necessary to generate the universally observed downward bending. In the simulation, new words' occurrence probability is predicted by Heaps' law and statistical fitting. According to Heaps' law, there is a good power-law relation between the size of a text (the number of word tokens) and the number of word types. Our statistical investigations into Heaps' exponents of 50 languages suggest that the distribution of Heaps' exponents is insensitive to genres and typological differences. The differential form of the function of heaps' law is the ratio of increase of word types to the increase of word tokens, i.e. the occurrence rates of new words. The Heaps' exponent is smaller than 1, and hence, the exponent of the corresponding differential form is smaller than 0, or, decreases with the increase of sample sizes. In other words, new words do not grow at a steady rate, but slower and slower, with increasing sample size.

Presently there are two influential models to account for the frequency distribution of words, both based on a single generating process. One is Simon model (*12, 26*), which assigns no fixed probabilities to high-frequency words, prescribing that new words appear with equal probabilities. This model has been attracting scholars because it can generate power law distributions of words and have explanatory power outside of linguistics. The other one is random typing model, which generates word frequency distribution by randomly hitting the keys of typewriters and calculating the frequencies of strings separated by spaces. This model, though mathematically simple, is capable of generating power law distributions. Figure 3 illustrates the simulation results of these two models. As shown in Figure 3, both models produce frequency-rank relations that, though in agreement with power-law functions, lack the downward bending found in natural languages (*27*). In other words, the frequency distribution of words in natural languages can be explained by neither of models that are respectively based on only one generating process. However, the proposed model based on the mechanism of dual-process yields curves with downward bending similar to those found in natural languages.

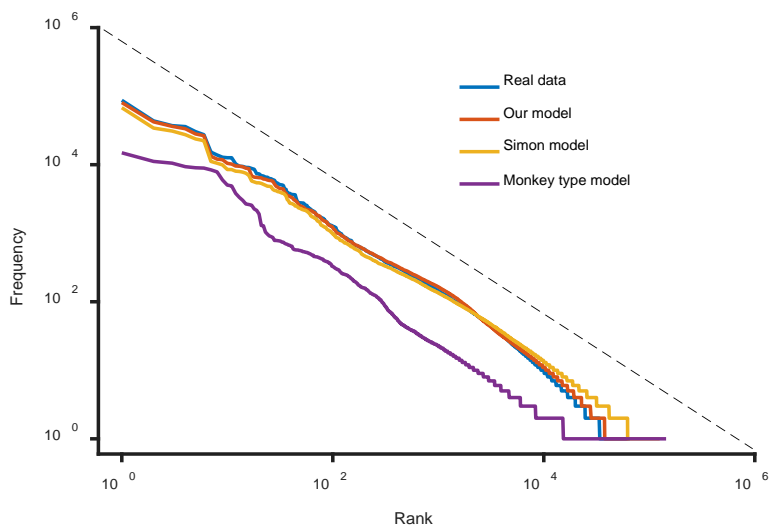

Figure 3. The frequency distributions generated by Simon model, random-typing model, and the model of our study.

## Discussion

The statistical results suggest that Zipf's curves of natural languages seem to universally have a 3-segment structural pattern: the upper segment, the middle segment, and the lower segment, as persistently found in 50 languages. Since each segment has its distinctive linguistic properties, Zipf's curve with this peculiar 3-segment structural pattern is likely to reflect a linguistic law, not a mere statistical artifact. In fact, Zipf and many other scholars have long noticed that Zipf's curve may consist of segments with different properties, as reflected in the observed deviation from expected curves (*28-29*). Some other scholars have tried two-segment piecewise fitting in English and obtained results that are capable of linguistic interpretations (*13, 30*). But their two-segment piecewise fittings are all based on one language sample, and hence cannot universally reveal the dynamic properties of these different segments. Based on 50 different languages, our findings probably suggest universal linguistic patterns that are underpinned by universal motivations such as common human cognition.

Since the 3 segments present different linguistic properties, it is possible to linguistically account for how this pattern is shaped. The upper segment of the curve covers mostly function words—the fundamental syntactic means in sentence organization and comprehension. Therefore, their frequencies in one language are expected to be stable regardless of text genres and sample sizes. However, syntactic means (e.g. function words, inflection, and word order) may weigh differently in typologically different languages, which is a possible linguistic explanation for our findings that typological differences have influence on exponents 1. Exponent 2 seems sensitive to text genre. Linguistically, this may have much to do with basic categories or concepts (*31*) because the middle segment consists mainly of most frequent content words. Texts of different genres may cover very different fields and topics, which may rely differently on these basic categories (concepts) in semantic presentation. As a result, the content words denoting these concepts may account for different proportions of words in texts of different genres, or rather, their relative frequencies may vary with different text genres. In addition, typological differences also affect the exponent of the middle segment, which may imply that typologically different languages may somewhat differ in the use of basic categories. However, in one language, the basic categories and their roles in semantic presentation are rather stable in a specific field, and hence the relative frequencies of these words should be stable, regardless of the text size. This is probably a linguistic explanation for the finding that exponent 2 is sensitive to differences in both typology and genre, but insensitive to differences in corpus size. Used quite infrequently, and in many cases only once, the words covered in the lower segment of the curves are largely neither function words nor basic category content words. For these words, their relative frequencies are not stable, with the average decreasing with larger sample sizes. What is noticeable is that the limiting value of exponent 3, insensitive to text genres, sample sizes or typological differences, is likely to be a linguistic universal shared by different languages.

Universal linguistic patterns are believed to be driven by more fundamental mechanisms, such as those of human cognition. As a result, the above linguistic patterns as reflected by the 3-segment structural pattern may also be accountable in terms of common human cognitive mechanism, which may further explained why this law universally exists in various languages and in many other human related areas. .

High-level human cognitive activities are probably characterized by the dual-process model. Type-1 process features promptness, automaticity, effortlessness, freedom from conscious attention, etc.; Type-2 process features slowness, controlled attention, conscious effort, etc. (*32*). The upper and the middle segments of Zipf's curve mainly cover those high-frequency words, whose frequencies rise in proportion to increase in sample size. This dynamic property means that there is no need to introduce any temporal parameter or other constraints into formal probability models to account for their probability of occurrence. Cognitively, the freedom from extra constraints means little or no effort for activation of these high-frequency words. Such effortlessness and promptness characterize Type-1 process. In contrast, the lower segment mainly covers those low-frequency words, whose relative frequencies are temporally unstable, decreasing with the growth of sample sizes. Therefore, temporal parameter or other constraints are needed to account for variable relative frequencies of low-frequency words. For these words, the relation between frequency and number of word types (frequency spectral) is also a power law function (*1*, *33*). This can be considered as the 1/f noise, a phenomenon widely found in both natural and social worlds. The frequency of low-frequency words is subject to modeling through non-stationary and non-ergodic stochastic processes (*34-35*). Such extra constraints cognitively correspond to extra mental efforts to activate or process these words. This effort-demanding processing is typical of type-2 process.

In fact, numerous psychological studies have evinced that word frequency bears closely on nearly all psychological/cognitive processes related to words, including intelligibility (*36-37*), recognizability (*38*), pronunciation (*39*), memory retention (*40*), naming time, and semantic categorization time (*41*), which is called word frequency effect. That is, the processing of high-frequency words call for less cognitive/psychological cost than low-frequency words (*42-43*). These findings are consistent with the findings of this study that high-frequency words and low-frequency words have very different dynamic properties. The probable reason is that language systems may rely heavily on the former for efficient Type-1 process, and at the same time, limit the growth rate of the latter to moderately regulate Type-2 process.

Governed by the principle of least effort, human cognition should mold language in such ways that Type-1 process can be fully utilized. The organization and comprehension of sentences depend heavily on syntactic means such as function words, which may play key roles in triggering Type-1 process so as to automatize syntactic organization and parsing. Cognitively, the access of function words, thanks to their high frequency, is largely effortless and automatic. As a result, function words may contribute to automatic and quick syntactic parsing. More sentences mean more function words, and hence their relative frequencies in one language should be stable regardless of genres and sample sizes. However, different syntactic means (unction words, inflection, and word order) may weigh differently in typologically different languages, which means the overall frequencies of function words may bear on typology.

Nevertheless, sentence comprehension is more than syntactic parsing—it ultimately aims to build semantic representations. Cognitively, the basic categories play vital and fundamental roles in our knowledge of world and semantic representation, frequently used, and cognitively easy to access. That is, these words may also be subject to Type-1 process. In brief, their frequency should steadily grow in proportion to the increase of sample sizes. However, since different genres may rely, to different degrees, on basic categories, the frequencies of these words are subject to influence of different genres.

Most words covered in the lower segment of the curve denote those concepts less fundamental in the world knowledge and thus are used infrequently. It is, hence, inefficient to constantly store these words in mind as independent units—it would be more economical to provisionally assemble a low-frequency word when needed, an operation calling for conscious attention and efforts. In other words, these words are largely subject to Type-2 process. Despite their assumed high processing cost, these words will not significantly deteriorate the general efficiency of language processing because they account for merely a quite limited proportion of word tokens in a language sample.

In brief, it is Type-1 process that shapes the upper and the middle segments of Zipf's curve, and Type-2 process, the lower segment. If human beings share similar cognitive mechanism, such as dual-process model, and if human languages are largely driven by cognition, it may be predicted that the 3-segment structural should be found in various languages, which is what has been found in our work on 50 languages.

**Conclusions**

Through the first large-scale cross-language investigation, our work has found that Zipf's laws in 50 languages all share a 3-segment structural pattern, with the lower segment invariably bending downward to deviate from theoretical expectation, and each segment demonstrating distinctive linguistic properties and different biases in use. Further computer simulations suggest the fundamental cognitive mechanism of dual-processing as a deep level motivation for the 3-segment structure. That is, Zipf's law found in natural languages is probably the result of the general constraint of human cognition. These findings suggest that Zipf's laws found in human-related fields are likely to be human driven, having deep-level motivations such as general cognitive mechanisms, and that big-data analysis into languages may provide valuable means to uncover language regularities and to trace cognitive laws hidden behind them.

**Acknowledgments**  This work is partly supported by the National Social Science Foundation of China (Grant No. 11&ZD188).

(2016-10-26)